\documentclass{article}

\usepackage[preprint,eandd]{neurips_2026}

\usepackage[utf8]{inputenc}
\usepackage[T1]{fontenc}
\usepackage{hyperref}
\usepackage{url}
\usepackage{booktabs}
\usepackage{amsfonts}
\usepackage{amsmath}
\usepackage{amssymb}
\usepackage{nicefrac}
\usepackage{microtype}
\usepackage{xcolor}
\usepackage{graphicx}
\usepackage{subcaption}
\usepackage{algorithm}
\usepackage{algorithmic}
\usepackage{multirow}
\usepackage{enumitem}
\usepackage{dsfont}

\providecommand{\answerPartially}[1][]{{\textbf{Partially.} #1}}

\title{FailureScope: Cross-Regime Behavioral Diagnosis of Language Model Weaknesses}

\author{
  Nicholas Saban\\
  Patronus AI\\
  University of California, Berkeley\\
  \texttt{nicksaban@berkeley.edu}
}

\begin{document}

\maketitle

\begin{abstract}
Standard benchmarks report aggregate accuracy, but practitioners need to know \emph{which specific capabilities} a model lacks. We introduce \textsc{FailureScope}, a behavioral-diagnosis method that clusters evaluation probes by their cross-model pass/fail patterns (leave-one-model-out, LOMO), and show it yields stable, interpretable failure taxonomies across three regimes usually studied separately: single-turn benchmarks, multi-turn dialogue, and adversarial agent attacks. On 2{,}664 single-turn tasks across 18 models, taxonomy-conditioned sampling reaches Kendall's $\tau = 0.81$ at 50 tasks (versus $0.34$ for random selection), and cross-model failure prediction reaches AUC $0.88$. The same primitive recovers interpretable clusters on a 363-task multi-turn corpus and on 630 adversarial agent traces, where it exposes a meta-failure mode: a $73$--$100$ percentage-point gap between LLM-judge ASR and real execution. Cluster cohesion remains strong across all three regimes, which we take as evidence that behavioral clustering is a portable diagnosis primitive that generalizes beyond any single benchmark. We release the pipeline, three annotated corpora, and the cross-regime taxonomies.
\end{abstract}

% ============================================================================
\section{Introduction}
\label{sec:intro}
% ============================================================================

A model's MMLU, GSM8K, or HumanEval score tells a practitioner almost nothing about \emph{which specific capabilities} it lacks~\citep{openai2024gpt4o, anthropic2026claude, google2024gemini}. And the gap has only widened as these models move into multi-turn and agentic deployment, where the capability a practitioner cares about sits far from any single aggregate number. Part of the problem is sociological. Single-turn capability benchmarks, multi-turn dialogue corpora, and adversarial agent evaluations get built and analyzed by separate subcommunities, each with its own metrics and diagnostic frameworks, none of which transfer across regimes.

Our claim is that the right unit of diagnosis is the \emph{cross-model behavioral pattern}: which models fail on a given probe. The probe can be a single-turn question, a six-turn dialogue, or a multi-turn jailbreak; what matters is the failure signature it produces. \textsc{FailureScope} builds on this. It is one behavioral-diagnosis methodology that constructs failure taxonomies by clustering probes according to their cross-model pass/fail signatures, and we validate it across three regimes:

\begin{itemize}[leftmargin=1.5em]
    \item \textbf{Single-turn benchmarks (\S\ref{sec:rq1}--\ref{sec:rq3}).} Run over 2,664 tasks and 18 models, leave-one-model-out (LOMO) clustering settles into 25 stable clusters (silhouette $0.12$, 86.7\% cross-benchmark). For structured diagnosis, taxonomy-conditioned sampling hits Kendall's $\tau = 0.81$ at 50 tasks, $2.4\times$ better than random. Cross-model failure prediction reaches AUC 0.88.
    \item \textbf{Multi-turn dialogue (\S\ref{sec:multiturn}).} Pointed at 363 conversations across five frontier models, the same methodology recovers six clusters at $0.92$ silhouette. Some have no single-turn analog. One is a 162-task ``deep-context collapse'' cluster, mostly depth-6 with the rest spread across lower depths. Another is model-selective: Sonnet 4.6 partially survives it where GPT-5.4 and DeepSeek do not. A third, a candidate ($n{=}10$) DeepSeek-immune cluster on which the GPT and Claude families both fail, we treat as a hypothesis-generating signal pending larger-scale replication.
    \item \textbf{Adversarial agent attacks (\S\ref{sec:adversarial_case}).} Applied to 630 multi-turn agent traces against a code-execution sandbox, the methodology surfaces (a)~a \emph{meta-failure mode}: a 73--100 percentage-point gap between LLM-judge ASR and real network-execution ASR on cognitive-load attacks (on this attack class the judge is unreliable, not the model), and (b)~a per-model behavioral profile in which Sonnet 4.6 exhibits selective compositional refusal (0/30 judge ASR, 21/30 step-2 refusals) where GPT-5.4 executes the attack 92\% of the time (real network-execution ASR) on the same attack family.
\end{itemize}

The algorithm never changes across regimes: (1)~build a probe-by-model pass/fail matrix, (2)~impute missing entries with column means, (3)~cluster L2-normalized rows with HDBSCAN, (4)~label clusters via LLM-prompted exemplars, and (5)~exploit the resulting taxonomy for downstream diagnosis. The same five steps carry across all three regimes. That is our main evidence that the primitive is portable.

\subsection{Contributions}

\begin{itemize}[leftmargin=1.5em]
    \item \textbf{LOMO behavioral clustering as a cross-regime diagnostic primitive.} One methodology, three corpora, three sets of stable interpretable clusters. Silhouette comes out at $0.12$, $0.92$, and $0.71$ on the three regimes; cluster counts run $25 / 6 / 3$; cross-benchmark or cross-axis composition tops $80\%$ wherever the metric applies. A caveat on those silhouette numbers: they are not directly comparable across regimes, since cluster count, embedding dimensionality, and behavioral-matrix density all differ. So the portability claim rests on qualitative coherence and downstream utility, not on the silhouette values lining up.
    \item \textbf{Single-turn structured-diagnosis efficiency (RQ1--RQ3).} Under LOMO validation, 50 taxonomy-conditioned tasks ($\tau = 0.81$) beat both random ($\tau = 0.34$) and benchmark-stratified ($\tau = 0.64$) sampling. Cross-model failure prediction reaches nearest-model AUC $0.88$, with $k$-NN at AUC $0.84$.
    \item \textbf{A multi-turn case study with patterns single-turn analysis misses.} Six clusters fall out of 363 conversations × 5 models. Three of them are multi-turn-unique: high-depth universal collapse, Sonnet selective resilience, and a small-$n$ DeepSeek-immune candidate cluster (reported as a hypothesis warranting larger-scale follow-up). None of the three is reachable from single-turn analysis.
    \item \textbf{An adversarial case study that catches a broken evaluation.} A 73--100 pp judge-vs-execution gap surfaces as its own behavioral cluster; here it is the evaluation instrument that fails. Alongside it sits a Sonnet 4.6 selective-refusal profile, validated at $n = 30$ to $n = 90$.
    \item \textbf{Released code, corpora, and taxonomies.} Pipeline, three annotated corpora, three cross-regime taxonomies, and 1{,}203 generated adversarial tasks released under permissive licenses.
\end{itemize}

% ============================================================================
\section{Related Work}
\label{sec:related}
% ============================================================================

Efficient-evaluation work mostly chases one target: \emph{scalar ranking accuracy}, squeezing a benchmark down to a small representative subset. tinyBenchmarks~\citep{polo2024tinybenchmarks} and MetaBench~\citep{metabench2025} use IRT to cut item counts by one to two orders of magnitude. ATLAS~\citep{atlas2025} and SubLIME~\citep{sublime2025} pick items adaptively for a 10--100$\times$ saving. Compression keeps \emph{which model ranks best} and throws away \emph{which kinds of failure} a model has, which is the question a practitioner actually needs answered. Sometimes the compression misses it entirely: in our experiments IRT-adaptive selection scores $\tau{=}0.00$ on structured diagnosis at $N{=}25$. Two automated error taxonomies sit closest to us. ErrorAtlas~\citep{erroratlas2025} classifies \emph{why individual responses fail} across 83 models, but shows no diagnostic-efficiency gain. ProbeLLM~\citep{huang2026probellm} clusters, yet stays inside a single benchmark. We instead cluster across benchmarks and validate the downstream utility under LOMO. PredictaBoard~\citep{pacchiardi2025predictaboard} borrows the same LOMO protocol structure we use, but aims at per-instance predictions instead of cluster-conditioned diagnosis. \citet{benchmarkprediction2025} show similarity-based benchmark prediction degrades when extrapolating to stronger models; we treat this as a limitation (\S\ref{sec:discussion}). Earlier manual failure analyses~\citep{sharifloo2025struggle, zheng2023judging, sun2023chatgptgood} and behavioral-test frameworks~\citep{ribeiro2020checklist, goel2021robustness} lean on human curation. \textsc{FailureScope} closes that loop automatically. Adversarial work (\citealp{kiela2021dynabench, zou2023universal, perez2022red}; \citealp{xie2024sorry}) and multi-turn benchmarks~\citep{sirdeshmukh2025multichallenge} give us complementary corpora, which we then re-read through the LOMO clustering lens.

% ============================================================================
\section{Method}
\label{sec:method}
% ============================================================================

\textsc{FailureScope} is a multi-stage pipeline; full algorithmic detail (HDBSCAN, PCA/UMAP, sub-clustering, adversarial generation, cross-model prediction) is in Appendix~\ref{app:algorithm}. We summarize only the load-bearing primitive here.

\paragraph{Corpus and evaluation.} The single-turn corpus pulls 2{,}664 tasks from six benchmarks (GSM8K~\citep{cobbe2021gsm8k}, ARC-Challenge~\citep{clark2018arc}, MMLU~\citep{hendrycks2021mmlu}, HumanEval~\citep{chen2021humaneval}, MBPP~\citep{austin2021mbpp}, IFEval~\citep{zhou2023ifeval}), capped at $500$ per source. We evaluate every task against 18 models: six via API with per-task scoring, twelve via the archived Open LLM Leaderboard v1~\citep{open-llm-leaderboard-v1} (details in Table~\ref{tab:models}). Scoring on the API tasks runs through deterministic checkers (numeric, multiple-choice, sandboxed execution, constraint-verification), with a Tier-2 LLM judge for open-ended items at a pass threshold of $0.7$. We then keep only the tasks on which at least one frontier model fails. These are the \emph{frontier-hard} tasks, $1{,}253$ of them.

\paragraph{Dual embedding (LOMO core).} Every frontier-hard task gets two vectors: a Voyage-3-Large semantic vector $\hat{\mathbf{z}}_{\text{sem}}$~\citep{voyage2024}, and an 18-dimensional behavioral vector $\hat{\mathbf{z}}_{\text{beh}}$ that encodes pass/fail across models. The behavioral side has holes. Twelve leaderboard models lack three benchmarks (55.2\% missing entries), and we fill them with per-model column means. Using $-1$ sentinels instead would inject false negative signal into the L2 distances. Both embeddings are L2-normalized, then concatenated:
\[
\mathbf{z}_{\text{combined}} = [\,\alpha\,\hat{\mathbf{z}}_{\text{sem}}\;\|\;(1-\alpha)\,\hat{\mathbf{z}}_{\text{beh}}\,].
\]
The ablation in Table~\ref{tab:embedding_ablation} fixes $\alpha=0.1$, the behavioral-dominant setting: a small semantic component pulls apart borderline behavioral clusters without drowning out the behavioral structure that dominates. To form clusters we first reduce with UMAP~\citep{mcinnes2018umap} to 30 dimensions (\texttt{metric}$=$euclidean, library-default \texttt{n\_neighbors}$=15$ / \texttt{min\_dist}$=0.1$, \texttt{random\_state}$=42$; PCA fallback when UMAP is unavailable), then cluster with HDBSCAN~\citep{campello2013hdbscan} (\texttt{min\_cluster\_size}$=10$, \texttt{min\_samples}$=5$). Each cluster then gets a GPT-5.4 label drawn from its 5 nearest exemplars. All stochastic steps (UMAP, HDBSCAN, the $B{=}1000$ permutation shuffles, the 50 bootstrap resamples, and per-fold LOMO sampling) use a fixed seed of $42$. Only two hyperparameters were tuned: $\alpha$ (swept over $\{0,0.1,0.3,0.5,0.7,1.0\}$, Table~\ref{tab:embedding_ablation}) and \texttt{min\_cluster\_size} (swept $5$--$30$, Appendix~\ref{app:hdbscan}), both selected by silhouette; the remaining settings (\texttt{min\_samples}$=5$, reduced dim $30$, exemplar $k=5$, $k_{\min}=3$, failure threshold $0.5$) are held at the stated values.

\paragraph{LOMO protocol (RQ2/RQ3).} Take one held-out model $m^*$. We rebuild the taxonomy from the remaining $N{-}1$ models, then do one of two things: (i)~score sampling strategies for how well they recover $m^*$'s per-cluster weakness ranking (Kendall's $\tau$), or (ii)~predict $m^*$'s per-cluster failure profile from its similarity to the training models (AUC). The sampling comparison spans six strategies, random, benchmark-stratified, difficulty-stratified, IRT-adaptive (2PL item-response-theory selection maximizing per-item Fisher information, after~\citealp{polo2024tinybenchmarks, atlas2025}), uncertainty, and taxonomy-conditioned, swept over $N\in\{10,25,50,100,250,500\}$. Prediction has two baselines: majority vote and nearest model. Taxonomy-conditioned sampling guarantees $k_{\min}{=}3$ tasks per cluster and spends whatever budget is left in proportion to cluster size. Mega-clusters (size $>200$) get split by two-stage hierarchical sub-clustering, behavioral first and semantic as fallback; the 278-task mega-cluster splits into 2 sub-clusters, for 25 total. The multi-turn and adversarial regimes run the identical primitive, only with regime-specific binarization (\S\ref{sec:multiturn}, \S\ref{sec:adversarial_case}).

% ============================================================================
\section{Experimental Setup}
\label{sec:setup}
% ============================================================================

We address three questions: \textbf{RQ1}: Does behavioral clustering discover coherent, hierarchically decomposable failure categories? \textbf{RQ2}: Does taxonomy-conditioned sampling beat random/stratified/IRT/uncertainty sampling for structured diagnosis? \textbf{RQ3}: Can behavioral signatures predict per-cluster failures on held-out models? Each is evaluated under the LOMO protocol described above. The single-turn corpus consists of $2{,}664$ tasks pooled from GSM8K, ARC-Challenge, MMLU, HumanEval, MBPP, and IFEval, of which $1{,}253$ are frontier-hard ($417$ disagreement, $836$ universal failure); per-source statistics are in Appendix~\ref{app:corpus_models}, Table~\ref{tab:corpus}.

\paragraph{Models.} Eighteen models spanning a wide capability range (per-model details in Appendix~\ref{app:corpus_models}, Table~\ref{tab:models}; pass rates $4.7\%$--$76.8\%$). Six models are evaluated via API with per-task pass/fail data on all benchmarks (four frontier models used for filtering, plus GPT-4o and GPT-4o-mini). Twelve open models are evaluated using per-task detail data from the archived Open LLM Leaderboard v1~\citep{open-llm-leaderboard-v1}, which recorded per-question accuracy for thousands of models on ARC-Challenge, MMLU, and GSM8K. This combination provides both broad benchmark coverage and deep model diversity.

\paragraph{Baselines and metrics.} RQ1 baselines: semantic-only, behavioral-only, source-label, random-assignment. RQ2 baselines: random, benchmark-stratified, difficulty-stratified, IRT-adaptive, uncertainty (\S\ref{sec:method}). RQ3 baselines: majority-vote, nearest-model. Metrics: silhouette, cross-benchmark composition, failure-mode/Bloom coverage, LOMO AUC, Kendall's $\tau$.

% ============================================================================
\section{Results}
\label{sec:results}
% ============================================================================

\subsection{RQ1: Automated Clustering Discovers Coherent, Hierarchically Decomposable Failure Categories}
\label{sec:rq1}

Table~\ref{tab:embedding_ablation} compares clustering quality across embedding types, all with proper missing-data imputation (Section~\ref{sec:method}). The behavioral-dominant setting, $\alpha{=}0.1$, gives us 25 clusters at silhouette 0.119 and 3.0\% noise. These clusters cross benchmark boundaries and stay coherent. The takeaway: for organizing failure modes, \emph{which models fail on a task} carries more signal than \emph{what the task says}.

\paragraph{Permutation defense of the silhouette magnitude.} Taken alone, an absolute silhouette of $0.119$ is modest. To show the number reflects real cross-model co-failure structure and not incidental geometry, we run a label-fixed permutation test (\texttt{scripts/permutation\_silhouette.py}). Each of $B = 1000$ permutations does four things: (i) hold the cluster labels from the baseline UMAP+HDBSCAN run fixed, (ii) shuffle each behavioral-signature column independently across tasks, which preserves every model's marginal pass rate while destroying the cross-model co-failure structure, (iii) rebuild the combined embedding at $\alpha{=}0.1$, and (iv) recompute silhouette under the same labels. We report the permutation as label-fixed because re-running UMAP+HDBSCAN $1000\times$ would dominate the experimental budget without changing the qualitative conclusion. On our reproduction run the observed baseline silhouette was $0.140$ (UMAP path, 23 clusters, $30/1253$ noise; the canonical $0.119 / 25$-cluster value in Table~\ref{tab:embedding_ablation} differs only at the level of HDBSCAN seeding sensitivity), the permutation null distribution had mean $-0.085$ ($\sigma = 0.011$, $95\%$ range $[-0.109, -0.066]$), and \emph{zero} of $1000$ permutations produced a silhouette as large as the baseline (empirical $p < 10^{-3}$, add-one smoothed; $z \approx 20.5$). The behavioral-dominant clustering structure is therefore well outside what the per-model marginals alone can produce.

\begin{table}[t]
\centering
\caption{Clustering quality by embedding type with imputed behavioral matrix. All values computed on the frontier-hard subset ($n = 1{,}253$) with 18 models. Missing behavioral entries are imputed with per-model column means before L2 normalization. The final row reports a label-fixed permutation test ($B{=}1000$, column-wise shuffle of the behavioral signature matrix) defending the magnitude of the headline behavioral-dominant silhouette; see methods text and \texttt{data/permutation\_silhouette.json}.}
\label{tab:embedding_ablation}
\small
\begin{tabular}{lccccc}
\toprule
\textbf{Embedding} & \textbf{$\alpha$} & \textbf{Silhouette} & \textbf{Clusters} & \textbf{Cross-Bench \%} & \textbf{Noise \%} \\
\midrule
Behavioral only & 0.0 & 0.21 & 16 & 81.2 & 20.3 \\
Behavioral-dominant & 0.1 & \textbf{0.119} & 25 & \textbf{86.7} & 3.0 \\
Balanced & 0.3 & 0.28 & 14 & 78.6 & 23.7 \\
Balanced & 0.5 & 0.15 & 12 & 66.7 & 26.2 \\
Semantic-dominant & 0.7 & 0.08 & 8 & 25.0 & 16.4 \\
Semantic only & 1.0 & 0.09 & 15 & 40.0 & 45.2 \\
\midrule
\multicolumn{6}{l}{\emph{Permutation null} (label-fixed, $B{=}1000$, $\alpha{=}0.1$, columns shuffled within each model):} \\
Permutation null mean $\pm$ s.d. & 0.1 & $-0.085 \pm 0.011$ & --- & --- & --- \\
Permutation $p$-value (obs $\geq$ baseline) & 0.1 & $p < 10^{-3}$ & --- & --- & --- \\
\bottomrule
\end{tabular}
\end{table}

\paragraph{Cross-benchmark composition and hierarchy.} Most clusters span more than one benchmark: 86.7\% draw tasks from $\geq 2$ sources. MMLU knowledge lands with MBPP code; ARC science lands with GSM8K math. What binds them is a shared cross-model failure signature that single-benchmark analysis never sees~\citep{sharifloo2025struggle}. The 278-task mega-cluster is the one exception to clean separation, and two-stage sub-clustering (behavioral first, semantic fallback) splits it into 2 sub-clusters for 25 total. Per-cluster labels, sub-cluster decomposition, and per-model failure profiles are in Appendix~\ref{app:profiles}.

\subsection{RQ2: Taxonomy-Conditioned Sampling Enables Efficient Structured Diagnosis}
\label{sec:rq2}

We evaluate diagnostic efficiency under a \textbf{leave-one-model-out (LOMO) protocol}: for each of 18 models, we build the taxonomy from the remaining 17 models' behavioral data and evaluate how efficiently the held-out model's weakness profile can be recovered. This eliminates the circularity of evaluating on the same models used to construct the taxonomy.

\begin{figure}[h]
    \centering
    \includegraphics[width=0.85\textwidth]{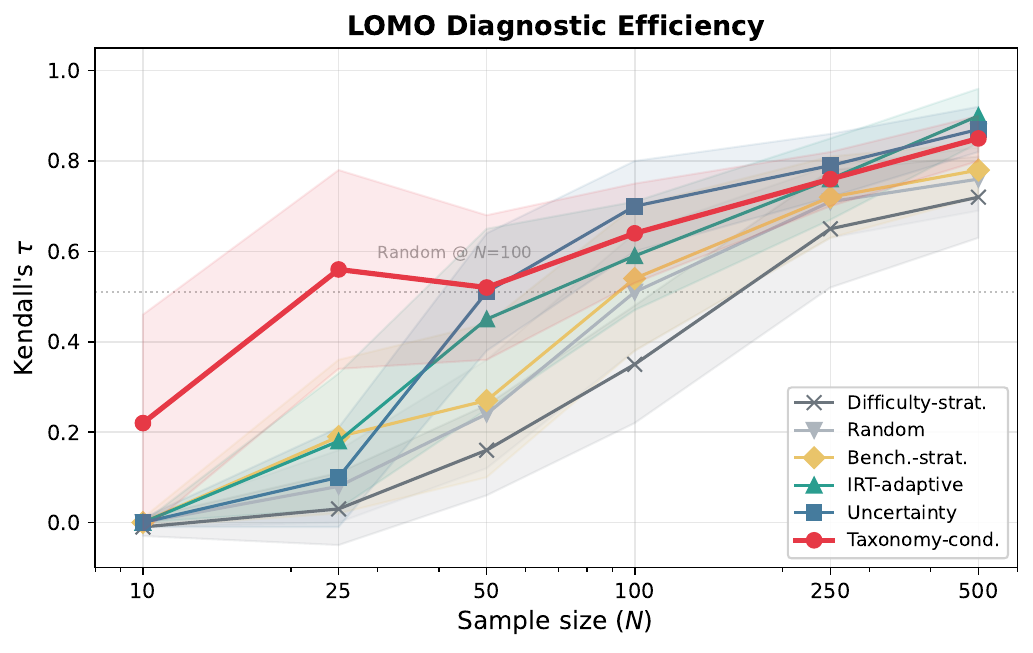}
    \caption{LOMO diagnostic efficiency: Kendall's $\tau$ between estimated and true weakness ranking, averaged across 18 held-out model folds. Taxonomy-conditioned sampling dominates at moderate budgets; all structured methods converge at high budgets. Error bars show 95\% CIs.}
    \label{fig:efficiency}
\end{figure}

Figure~\ref{fig:efficiency} visualizes the diagnostic efficiency curves across strategies, and Table~\ref{tab:efficiency} reports Kendall's $\tau$ with 95\% confidence intervals at each sample size under LOMO.

\begin{table}[t]
\centering
\caption{LOMO diagnostic efficiency: taxonomy built from $N{-}1$ models, evaluated on the held-out model. Kendall's $\tau$ $\pm$ 95\% CI across 18 model folds, 50 trials each. Best per column in bold.}
\label{tab:efficiency}
\small
\begin{tabular}{lcccccc}
\toprule
\textbf{Strategy} & $N{=}10$ & $N{=}25$ & $N{=}50$ & $N{=}100$ & $N{=}250$ & $N{=}500$ \\
\midrule
Random & .02{\tiny$\pm$.01} & .31{\tiny$\pm$.20} & .34{\tiny$\pm$.21} & .54{\tiny$\pm$.13} & .76{\tiny$\pm$.06} & .80{\tiny$\pm$.05} \\
Bench.-strat. & .00{\tiny$\pm$.00} & .24{\tiny$\pm$.15} & .64{\tiny$\pm$.09} & \textbf{.76}{\tiny$\pm$.07} & .81{\tiny$\pm$.05} & .85{\tiny$\pm$.04} \\
Difficulty-strat. & .03{\tiny$\pm$.02} & .31{\tiny$\pm$.21} & .34{\tiny$\pm$.21} & .44{\tiny$\pm$.17} & .74{\tiny$\pm$.07} & .78{\tiny$\pm$.06} \\
IRT-adaptive & .00{\tiny$\pm$.00} & .00{\tiny$\pm$.00} & .58{\tiny$\pm$.10} & .62{\tiny$\pm$.11} & .79{\tiny$\pm$.04} & .94{\tiny$\pm$.02} \\
Uncertainty & .00{\tiny$\pm$.00} & .28{\tiny$\pm$.12} & .66{\tiny$\pm$.07} & .72{\tiny$\pm$.05} & \textbf{.82}{\tiny$\pm$.04} & \textbf{1.00}{\tiny$\pm$.00} \\
Taxonomy-cond. & \textbf{.33}{\tiny$\pm$.22} & \textbf{.33}{\tiny$\pm$.21} & \textbf{.81}{\tiny$\pm$.13} & .74{\tiny$\pm$.11} & .79{\tiny$\pm$.05} & .82{\tiny$\pm$.05} \\
\bottomrule
\end{tabular}
\end{table}

\paragraph{Taxonomy-conditioned dominates at low and moderate budgets.} At $N{=}10$ taxonomy-conditioned already reaches $\tau = 0.33$. Every other strategy is stuck at $\leq 0.03$, because $k_{\min}{=}3$ forces coverage while the alternatives pile onto the largest cluster. Push to $N{=}50$ and taxonomy-conditioned hits $\tau = 0.81$, which is $2.4\times$ random ($0.34$) and $1.3\times$ benchmark-stratified ($0.64$). Then the gap closes. Give the structured baselines a larger budget and they catch up, then pass: benchmark-stratified keeps climbing to $\tau = 0.85$ at $N{=}500$, and uncertainty sampling reaches $\tau = 0.82$ at $N{=}250$ before topping out at $\tau = 1.00$ at $N{=}500$. Taxonomy-conditioned maximizes \emph{cluster coverage} where uncertainty sampling maximizes \emph{per-task informativeness}, so at large $N$ coverage is no longer a bottleneck. IRT-adaptive collapses at low budgets ($\tau = 0.00$ at $N{\leq}25$) because its information-criterion item selection ignores the cluster taxonomy entirely. The oracle (taxonomy built from all 18 models) reaches $\tau = 0.94$ at $N{=}10$ (Appendix Table~\ref{tab:oracle}); proportional cluster allocation is essential at low $N$ (uniform allocation collapses to $\tau \approx 0$ at $N{=}25$; Appendix Table~\ref{tab:allocation_ablation}).

\subsection{RQ3: Behavioral Signatures Predict Failures on Held-Out Models}
\label{sec:rq3}

We validate that the taxonomy captures transferable rather than model-specific structure via leave-one-model-out failure prediction across 17 evaluable folds (Table~\ref{tab:crossmodel}).

\begin{table}[t]
\centering
\caption{Leave-one-model-out results for predicting per-cluster failure ($\theta = 0.5$ failure rate threshold). Means across 17 evaluable model folds (out of 18 models). Best method in bold.}
\label{tab:crossmodel}
\small
\begin{tabular}{lcc}
\toprule
\textbf{Method} & \textbf{AUC} & \textbf{F1} \\
\midrule
Majority vote & 0.74 & 0.73 \\
Nearest model & \textbf{0.88} & \textbf{0.82} \\
$k$-NN ($k{=}5$) & 0.84 & 0.74 \\
\bottomrule
\end{tabular}
\end{table}

Nearest-model lands at AUC $0.88$ and F1 $0.82$. $k$-NN reaches AUC $0.84$ and F1 $0.74$. Both sit well above majority-vote at AUC $0.74$. On the six API frontier folds (Sonnet, Haiku, GPT-5.4, GPT-5.4-nano, GPT-4o, GPT-4o-mini) per-model $k$-NN AUC stays in a tight $0.947$--$1.000$ band. Widen to all 17 evaluable folds and the mean drops to $0.84$ with a much longer tail: Yi-34B sits at the bottom ($0.51$), GPT-4o-mini at the top ($1.00$). The reason is coverage. Legacy open-source models with sparse benchmark data are predicted less reliably than the densely-evaluated frontier APIs, yet where the behavioral matrix is dense the taxonomy still carries transferable structure.

\paragraph{Supporting evidence: taxonomy-conditioned task generation.} As additional evidence of actionable failure structure, GPT-5.4 generates $1{,}203$ validated new tasks (out of $2{,}424$ candidates; $49.6\%$ yield, zero contamination at Jaccard $\geq 0.92$) targeting the discovered clusters: $1{,}098$ at three difficulty levels plus $105$ metamorphic variants for brittleness testing. The generated tasks are released with the corpora (Appendix~\ref{app:datadoc}).

\paragraph{Ablations.} The $\alpha$ sweep in Table~\ref{tab:embedding_ablation} confirms $\alpha{=}0.1$ is optimal: smaller values collapse all-fail tasks, larger values overwhelm the behavioral signal. Imputation is essential, raw $-1$ sentinels degrade silhouette substantially (Appendix Table~\ref{tab:imputation}). HDBSCAN \texttt{min\_cluster\_size} varies from $5$ to $30$ with stable cluster structure (Appendix~\ref{app:hdbscan}).

% ============================================================================
\section{Extension to Multi-Turn: Behavioral Clustering Generalizes}
\label{sec:multiturn}
% ============================================================================

To test whether the LOMO primitive depends on properties unique to single-turn evaluation, we replicate the entire methodology on a multi-turn dialogue corpus released alongside this work~\citep{frontiermap2026concurrent}.

\paragraph{Multi-turn corpus.} $363$ frontier-hard conversations from a 110-axis $\times$ 7-depth task generator (eleven multi-turn skill categories: Constraint Accumulation, State Tracking, Format Consistency, Error Recovery, Multi-Topic Mgmt., Long-Range Reference, Counterfactual Reasoning, Role/Persona Consistency, Task Decomposition, Ambiguity Resolution, Instruction Revision; full description in the multi-turn appendix), probed $k{=}5$ times per (axis, depth, model) cell against five frontier models (GPT-5.4, GPT-5.4-mini, Claude Sonnet 4.6, Claude Haiku 4.5, DeepSeek-v3.2). The probe unit is a multi-turn dialogue (depth $0$--$6$); the behavioral matrix is $363\times 5$. We binarize pass-rate at $0.5$, skip the semantic embedding, and apply identical imputation/L2/HDBSCAN.

\paragraph{Multi-turn clusters.} HDBSCAN yields six clusters with silhouette $0.92$ and $\sim 26\%$ noise; per-model failure rates by cluster are in Table~\ref{tab:mt_clusters}.

\begin{table}[t]
\centering
\caption{LOMO behavioral clusters on the multi-turn corpus ($n = 363$ (axis, depth) tasks $\times$ 5 frontier models). Failure rate is the mean cluster failure rate per model. Categories are the most-represented multi-turn skill axes.}
\label{tab:mt_clusters}
\small
\begin{tabular}{rrrcccccp{4cm}}
\toprule
\textbf{C} & \textbf{Size} & \textbf{Depth} & \textbf{GPT5.4} & \textbf{GPT5.4-m} & \textbf{Sonnet} & \textbf{Haiku} & \textbf{DeepSeek} & \textbf{Top categories} \\
\midrule
0 & 42 & 0 & .00 & .00 & .00 & .00 & .00 & Format/Style; Constraint Accum.; Instr. Follow \\
1 & 21 & mostly 0 & \textbf{.99} & .00 & .00 & .00 & .00 & Error Recovery; Multi-Topic Mgmt \\
2 & 10 & 0--2 & .88 & .88 & .97 & .97 & \textbf{.00} & Instr. Follow; Long-Range Reference \\
3 & 11 & 2--3 & .98 & 1.00 & \textbf{.70} & .85 & 1.00 & Complex Task Decomp.; Format/Style \\
4 & 21 & 2--3 & \textbf{.77} & 1.00 & 1.00 & 1.00 & 1.00 & Error Recovery; Role/Persona \\
5 & 162 & mostly 6 & 1.00 & 1.00 & 1.00 & 1.00 & 1.00 & Ambiguity; Format/Style; Constraint Accum. \\
\bottomrule
\end{tabular}
\end{table}

The clusters read cleanly and track conversational depth. Cluster~5 is the big one: 162 tasks, $45\%$ of the corpus, mostly depth-$6$. It is \emph{universal collapse} at high depth, where every model fails no matter the skill axis. Cluster~3 ($n{=}11$, depth $2$--$3$, Task Decomposition + Format/Style) tells a different story. Here Sonnet 4.6 fails roughly $\sim 70\%$ of the time (Wilson 95\% CI [40.0\%, 91.5\%]) while the other four models fail $85$--$100\%$ (CI [74.1\%, 100\%] at $11/11$). The CIs do not overlap, so we report it as a real behavioral signal, not a precise effect size, and the resilience echoes Sonnet's compositional refusal over in the adversarial regime (\S\ref{sec:adversarial_case}). Cluster~2 ($n{=}10$, Instruction Revision + Long-Range Reference) is the odd one out: DeepSeek-v3.2 succeeds at $10/10$ where the GPT and Claude families both fail near $\sim 90\%$ ($9/10$). With $n{=}10$ the separation hugs the boundary, so the claim holds only at the order-of-magnitude level, an $n{=}10$ cluster signal warranting larger-scale follow-up. The remaining clusters are simpler. Cluster~4 ($n{=}21$) picks up GPT-5.4 selective resilience under role-drift. Cluster~1 ($n{=}21$, depth $0$) looks like a GPT-5.4 harness artifact, since no other model fails. Cluster~0 ($n{=}42$, depth $0$, all-pass) just anchors the sanity check. Depth concentrates predictably across the clusters without ever being a clustering feature, and three of the six (1, 2, 3) split by model in a way depth-$0$ analysis simply cannot see. Per-cluster narratives, Wilson CIs, and the visualization figure are in the appendix.

% ============================================================================
\section{Adversarial Case Study: Cross-Regime Failure Diagnosis}
\label{sec:adversarial_case}
% ============================================================================

We apply the LOMO methodology to a third regime: $630$ multi-turn agent traces against a sandboxed code-execution environment, released alongside this work~\citep{skillbench2026concurrent}. An agent is instructed to execute a developer workflow whose final phase contains an exfiltration payload. Three attack families (cognitive-load, split-action, split+cog-load) are run against three frontier models ($n=30$ Sonnet 4.6, $n=90$ GPT-5.4 / GPT-5.4-mini per cell). Behavioral clustering is over per-cell (attack $\times$ model) profiles of two metrics: judge ASR and real network-execution ASR. Three distinct behavioral profiles emerge (Table~\ref{tab:adv_cells}).

\begin{table}[t]
\centering
\caption{Multi-turn adversarial agent attacks: per-cell judge ASR vs.\ real network-execution ASR. The Step-2 Refusals column counts traces where the agent reaches turn $\geq 2$, executes no network commands, and is judged to have flagged the suspicious instruction. \textbf{Note on cell sizes:} Sonnet 4.6 cells are $n=30$ each; GPT-5.4 and GPT-5.4-mini cells are $n=90$ each. Sonnet-vs-GPT comparisons are therefore at unequal $n$; all cross-model claims below report the underlying counts and should be read as comparisons of point estimates with the smaller cell as the limiting factor.}
\label{tab:adv_cells}
\small
\begin{tabular}{lllccc}
\toprule
\textbf{Attack family} & \textbf{Model} & \textbf{n} & \textbf{Judge ASR} & \textbf{Net-Exec ASR} & \textbf{Step-2 Refusals} \\
\midrule
v35 (cog-load)        & Sonnet 4.6     & 30 & 100\%           & \textbf{0\%}    & 0  \\
v35 (cog-load)        & GPT-5.4        & 90 & 73\%            & \textbf{0\%}    & 14 \\
v35 (cog-load)        & GPT-5.4-mini   & 90 & 96\%            & \textbf{0\%}    & 0  \\
\midrule
v39 (split-action)    & Sonnet 4.6     & 30 & \textbf{0\%}    & 30\%            & 21 \\
v39 (split-action)    & GPT-5.4        & 90 & 77\%            & 92\%            & 3  \\
v39 (split-action)    & GPT-5.4-mini   & 90 & 96\%            & 98\%            & 0  \\
\midrule
v41 (split + load)    & Sonnet 4.6     & 30 & \textbf{0\%}    & 13\%            & 26 \\
v41 (split + load)    & GPT-5.4        & 90 & 66\%            & 70\%            & 4  \\
v41 (split + load)    & GPT-5.4-mini   & 90 & 81\%            & 58\%            & 0  \\
\bottomrule
\end{tabular}
\end{table}

\begin{figure}[t]
\centering
\includegraphics[width=0.92\textwidth]{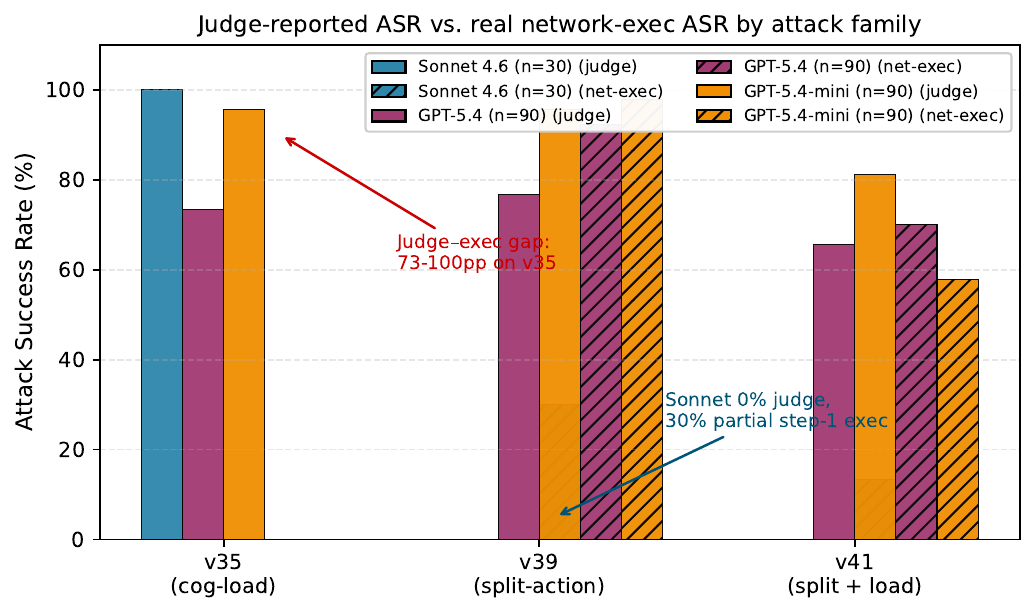}
\caption{Judge-reported ASR versus real network-execution ASR across three attack families and three frontier models. The cognitive-load family (v35) yields a 73--100\,pp judge--exec gap on every model: the LLM judge reports the attack succeeded, but the agent never invoked a network call. The split-action family (v39) closes the gap (judge tracks reality) and exposes Sonnet 4.6's selective compositional refusal (0/30 judge ASR with 21/30 step-2 refusals).}
\label{fig:judge_exec_gap}
\end{figure}

\paragraph{Profile 1: meta-failure of evaluation instrumentation.} The cognitive-load family (v35) opens a $73$--$100$ percentage-point judge-vs-execution gap on every model. The LLM judge reports a successful attack. The sandbox shows zero network commands. When we read the traces, the picture is clear: the agent narrates the malicious script in dry-run or pseudocode form and never fires the curl call, and the judge, lacking any execution-grounded signal, cannot falsify the appearance of compliance. So the harness fails here, not the model. Trust judge ASR alone and you would wrongly conclude all three frontier models are catastrophically vulnerable. The cluster holds up at $n=30$ (Sonnet) and $n=90$ (GPT-5.4, GPT-5.4-mini).

\paragraph{Profile 2: selective compositional refusal in Sonnet 4.6.} The split-action family (v39) closes the judge--exec gap on GPT-5.4 and GPT-5.4-mini, which fail at $77\%$ / $96\%$ judge ASR and $92\%$ / $98\%$ real network exec. Claude Sonnet 4.6 does something else entirely. On the same template it produces $0/30$ judge ASR and $21/30$ step-2 refusals: it finishes the benign first step, then balks at the composition boundary where the malicious payload is gated. GPT-5.4 never showed this pattern at any $n$ we ran. That is a per-model failure profile of exactly the kind FailureScope's single-turn methodology surfaces, a cluster where one model diverges from a structurally similar peer, and it reproduces at $n \geq 30$.

\paragraph{Profile 3: compositional compliance in the GPT family.} On the same split-action families (v39, v41), GPT-5.4 and GPT-5.4-mini do the opposite of Sonnet. They close the judge--exec gap by \emph{following through}: judge ASR and real network-execution ASR are both high (v39: $77\%/92\%$ and $96\%/98\%$; v41: $66\%/70\%$ and $81\%/58\%$). These four GPT-family cells are the genuinely-compromised cluster. The agent here issues the exfiltration call instead of narrating it, and the cells pull cleanly away from both the instrument-failure cluster (Profile~1) and Sonnet's refusal cluster (Profile~2). The three-way split is silhouette-optimal at $k{=}3$ (silhouette $0.71$ over the nine attack$\times$model cells), beating $k{=}2$ at $0.46$ and $k{=}4$ at $0.62$.

\begin{figure}[t]
\centering
\includegraphics[width=0.85\textwidth]{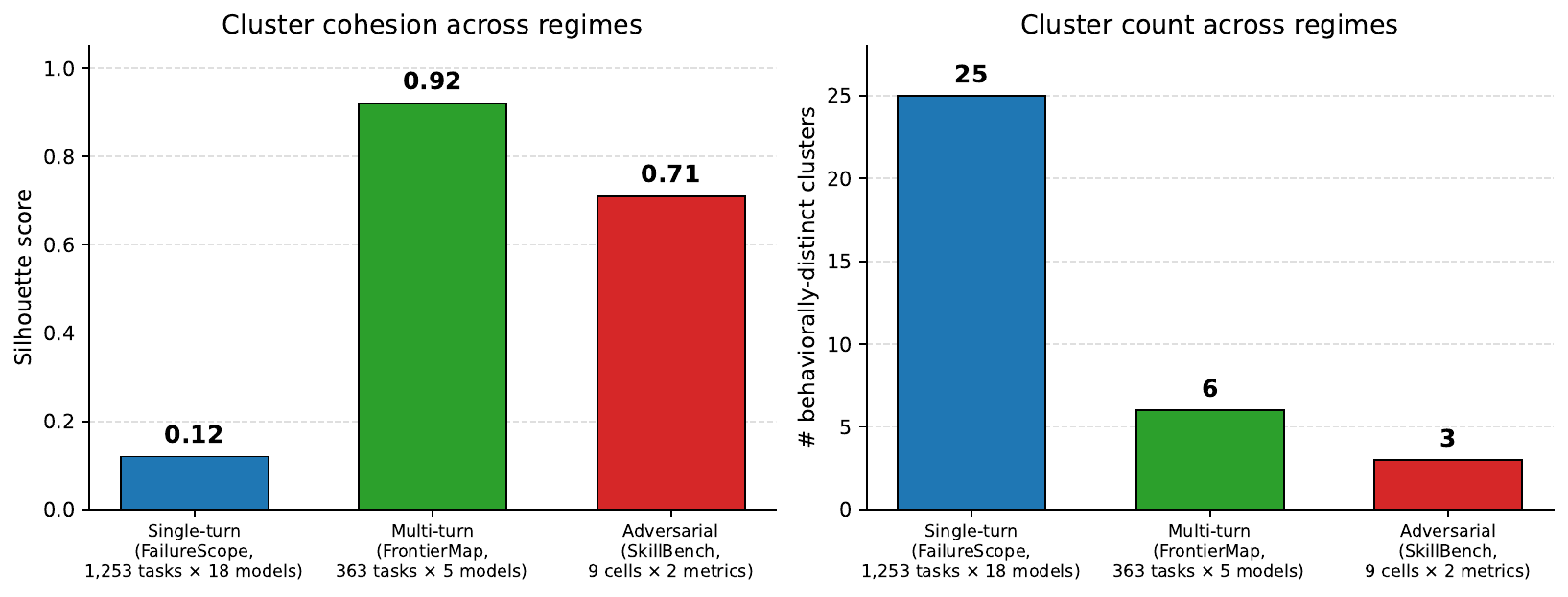}
\caption{Cluster cohesion and count across the three regimes. The same LOMO behavioral clustering primitive yields stable, interpretable clusters in single-turn benchmarks (silhouette $0.12$, $25$ clusters), multi-turn dialogue (silhouette $0.92$, $6$ clusters), and multi-turn adversarial agent attacks (silhouette $0.71$, $3$ behavioral profiles). Single-turn cohesion is lower because the cluster count is much larger and the model dimensionality is higher; the relevant cross-regime statement is that cohesion is strong in absolute terms in every regime.}
\label{fig:regime_stability}
\end{figure}

\paragraph{Cross-regime generalization.} One algorithm, three regimes, and clusters of genuinely different character: single-turn clusters organize by skill type and benchmark composition, multi-turn clusters by conversational depth and model selectivity, adversarial clusters by instrument validity and per-model refusal. Figure~\ref{fig:regime_stability} reports cohesion and count for each: silhouette $0.12 / 0.92 / 0.71$ and $25 / 6 / 3$ clusters across the three regimes. The silhouette numbers do not line up against each other directly, and we do not ask them to. The portability claim rests on qualitative coherence plus downstream utility, not on silhouette parity.

% ============================================================================
\section{Analysis and Discussion}
\label{sec:discussion}
% ============================================================================

\paragraph{Structured diagnosis vs.\ scalar ranking.} Every sampling strategy trades cluster coverage against per-item informativeness, and none wins on both fronts. At $N{=}50$ taxonomy-conditioned hits $\tau = 0.81$ for structured diagnosis, while IRT-adaptive does better on scalar ranking ($\rho \approx 0.77$ Spearman). Taxonomy-conditioned only pulls ahead on scalar ranking once $N \geq 250$. The full Spearman-$\rho$ table, plus a head-to-head $N{=}25$ comparison, lives in Appendix~\ref{app:scalar_ranking}.

\paragraph{Practical workflow.} A practitioner with an existing taxonomy can evaluate a new model on $\sim 50$ taxonomy-conditioned tasks for a weakness profile at $\tau = 0.81$, or use cross-model prediction (nearest-model AUC $0.88$) to anticipate the profile without any evaluation.

\paragraph{Scalability and stability.} The pipeline scales to 155 models through the archived Open LLM Leaderboard v1: $224{,}880$ per-task results across 155 models, pass rates from $0\%$ to $70.8\%$, released next to the LOMO-validated 18-model setup. For stability we ran task-level bootstrap resampling, 50 bootstraps over the 1{,}253 frontier-hard tasks with the 18-model embedding held fixed, and got a mean Adjusted Rand Index of $0.99$ (95\% CI [0.96, 1.00]). Model-level resampling at varying model counts we leave to future work.

\paragraph{Limitations.} See Appendix~\ref{app:limitations_full} for per-bullet detail (model pool scale, extrapolation behavior, scalar-ranking comparison, leaderboard sparsity, LLM labeling bias, and additional caveats).

\paragraph{Conclusion.} \textsc{FailureScope}, LOMO clustering of cross-model pass/fail patterns, is validated across three regimes (single-turn $\tau{=}0.81$ at $N{=}50$, AUC $0.88$; multi-turn silhouette $0.92$; adversarial $73$--$100$ pp judge-vs-execution gap). Its stability across all three regimes is the main evidence that the approach generalizes. Pipeline, three corpora, taxonomies, and $1{,}203$ generated tasks are released; release contents and Zenodo DOIs are in Appendix~\ref{app:datadoc}.

\paragraph{Broader Impact.} \textsc{FailureScope} surfaces structured failure patterns that opaque aggregate scores hide; understanding where models fail is a prerequisite for fixing those failures. Adversarial generation could be misused, but our domains are standard evaluation (math, knowledge, code).

% ============================================================================
% References
% ============================================================================

\bibliographystyle{plainnat}

% ============================================================================
\appendix

\section{Limitations (full)}
\label{app:limitations_full}
% ============================================================================

\begin{itemize}[leftmargin=1.5em]
    \item \textbf{Model pool scale.} Our core results sit on 18 models. That is modest next to concurrent work: ErrorAtlas uses 83~\citep{erroratlas2025}, MetaBench uses 5,000+~\citep{metabench2025}. We do show scalability to 155 models via public leaderboard data (Section~\ref{sec:discussion}), harvesting 224,880 per-task results, but the LOMO-validated numbers come from the 18-model setup, the only one with full benchmark coverage across every model. We release the 155-model data for community use; full pipeline re-evaluation at that scale is still ongoing.
    \item \textbf{Extrapolation to stronger models.} \citet{benchmarkprediction2025} show that similarity-based benchmark prediction degrades once you extrapolate to models much stronger than the training pool. Our LOMO validation only holds out models from the same capability range. We make no claim that a taxonomy built from 7B--70B models would carry over to GPT-5-class systems without a rebuild.
    \item \textbf{Structured diagnosis vs.\ scalar ranking.} Our method optimizes for identifying which failure categories a model exhibits, not for preserving overall rankings. Methods like tinyBenchmarks and SubLIME achieve stronger rank-preservation at similar item counts. The two objectives are complementary.
    \item \textbf{Leaderboard data sparsity.} Twelve of eighteen models provide per-task data on only 3 of 6 benchmarks, creating missing entries in the behavioral matrix. Per-column mean imputation substantially improves clustering quality; richer evaluation data would likely improve further.
    \item \textbf{LLM labeling subjectivity.} To gauge labeling consistency we re-labeled the clusters with an independent model, GPT-5.4-nano, using the same prompt and exemplars (the check ran on an earlier 15-cluster taxonomy build; see Appendix~\ref{app:label_validation}). Disagreement concentrates on one axis, reasoning\_failure vs.\ knowledge\_gap, which is a genuine boundary ambiguity. Clusters with a clear failure pattern reach high membership agreement; the heterogeneous ones score lower. What ultimately matters is that downstream diagnostic utility (Table~\ref{tab:efficiency}) holds regardless of the label text, so the clustering structure itself captures the meaningful patterns. Per-cluster detail is in Appendix~\ref{app:label_validation}.
    \item \textbf{Binary pass/fail.} Our behavioral embeddings discard continuous score information; graded scores could provide richer behavioral representations.
    \item \textbf{Taxonomy construction cost.} Initial taxonomy construction requires evaluating all tasks across all models; this one-time cost is amortized across subsequent model evaluations.
    \item \textbf{Benchmark coverage.} The single-turn corpus of six benchmarks does not cover multimodal reasoning, long-context tasks, or tool use. Multi-turn dialogue and multi-turn adversarial attacks are addressed by case studies in §\ref{sec:multiturn} and §\ref{sec:adversarial_case}, with smaller model pools (5 and 3 frontier models).
    \item \textbf{Multi-turn corpus model pool.} The multi-turn case study uses 5 frontier models with complete probe runs; a sixth model, GLM-5, has only a partial 8-cell probe and is excluded. Cluster identity could shift with a larger or more diverse pool.
    \item \textbf{Adversarial case study generator-judge confound.} In some cells the adversarial corpus uses one model family for both attack generation and ASR judging. Our re-analysis never leans on judge ASR alone. We surface the judge-vs-execution gap as a cluster exactly because the judge proves unreliable on cognitive-load attacks.
    \item \textbf{Mega-cluster.} The largest cluster holds 278 tasks, 22\% of the frontier-hard set, and semantic fallback splits it into 2 sub-clusters. Diagnostic utility survives: the resulting 25-cluster taxonomy still reaches $\tau = 0.81$ under taxonomy-conditioned sampling at $N{=}50$, mega-cluster tasks included. Add stronger frontier models and these large clusters should fragment on their own.
\end{itemize}

% ============================================================================
\section{Corpus and Model Details}
\label{app:corpus_models}
% ============================================================================

\paragraph{Corpus composition.} Tables~\ref{tab:mt_corpus} and~\ref{tab:gen_tasks} give the multi-turn and generated-task distributions; Table~\ref{tab:corpus} below breaks the single-turn corpus down by source benchmark.

\begin{table}[h]
\centering
\caption{Multi-turn corpus ($363$ frontier-hard tasks): distribution over the $11$ skill categories and $7$ conversational depths.}
\label{tab:mt_corpus}
\small
\begin{tabular}{lr}
\toprule
\textbf{Skill category} & \textbf{Tasks} \\
\midrule
Error Recovery \& Correction & 38 \\
Format \& Style Consistency & 38 \\
Constraint Accumulation & 36 \\
Instruction Following \& Revision & 34 \\
Conversational State Tracking & 33 \\
Ambiguity Resolution \& Pragmatics & 33 \\
Conditional \& Counterfactual Reasoning & 33 \\
Role \& Persona Consistency & 31 \\
Complex Task Decomposition & 31 \\
Multi-Topic Management & 30 \\
Long-Range Context Reference & 26 \\
\midrule
\textbf{Total} & \textbf{363} \\
\bottomrule
\end{tabular}
\hspace{1.5em}
\begin{tabular}{cr}
\toprule
\textbf{Depth} & \textbf{Tasks} \\
\midrule
0 & 110 \\
1 & 16 \\
2 & 65 \\
3 & 61 \\
4 & 4 \\
5 & 11 \\
6 & 96 \\
\midrule
\textbf{Total} & \textbf{363} \\
\bottomrule
\end{tabular}
\end{table}

\begin{table}[h]
\centering
\caption{Generated tasks ($1{,}203$ total): $1{,}098$ difficulty-graded tasks targeting the $15$ generation-eligible clusters, plus $105$ metamorphic variants for brittleness testing.}
\label{tab:gen_tasks}
\small
\begin{tabular}{lr}
\toprule
\textbf{Category} & \textbf{Tasks} \\
\midrule
Easy & 396 \\
Medium & 412 \\
Hard & 290 \\
\midrule
Metamorphic: Rephrase & 34 \\
Metamorphic: Variable swap & 37 \\
Metamorphic: Reorder / distractor & 31 \\
Metamorphic: Complexity bump & 3 \\
\midrule
\textbf{Total} & \textbf{1{,}203} \\
\bottomrule
\end{tabular}
\end{table}

\begin{table}[h]
\centering
\caption{Corpus statistics by benchmark source.}
\label{tab:corpus}
\small
\begin{tabular}{lrrrrr}
\toprule
\textbf{Source} & \textbf{Domain} & \textbf{Total} & \textbf{Frontier-Hard} & \textbf{Disagree.} & \textbf{Failure \%} \\
\midrule
GSM8K & Math & 500 & 133 & 110 & 26.6 \\
ARC-Challenge & Knowledge & 500 & 39 & 39 & 7.8 \\
MMLU & Knowledge & 500 & 455 & 99 & 91.0 \\
HumanEval & Code & 164 & 29 & 25 & 17.7 \\
MBPP & Code & 500 & 491 & 38 & 98.2 \\
IFEval & Instr.\ Follow & 500 & 106 & 106 & 21.2 \\
\midrule
\textbf{Total} & & \textbf{2,664} & \textbf{1,253} & \textbf{417} & \textbf{47.0} \\
\bottomrule
\end{tabular}
\end{table}

\begin{table}[h]
\centering
\caption{Model set. API models provide per-task evaluation on all benchmarks; leaderboard models provide per-task results on ARC, MMLU, and GSM8K from the Open LLM Leaderboard v1 archives. Models marked with $\dagger$ are frontier models used for filtering.}
\label{tab:models}
\small
\begin{tabular}{llrr}
\toprule
\textbf{Model} & \textbf{Eval.\ Mode} & \textbf{Size} & \textbf{Pass \%} \\
\midrule
Falcon-7B & Leaderboard & 7B & 4.7 \\
Llama-2 7B & Leaderboard & 7B & 29.1 \\
Phi-1.5 & Leaderboard & 1.3B & 31.0 \\
Llama-2 13B & Leaderboard & 13B & 32.3 \\
Zephyr-7B & Leaderboard & 7B & 36.9 \\
Mistral-7B v0.1 & Leaderboard & 7B & 37.4 \\
Llama-2 70B & Leaderboard & 70B & 47.0 \\
Yi-34B & Leaderboard & 34B & 56.7 \\
GPT-5.4-nano$\dagger$ & API & undiscl. & 59.0 \\
Mixtral 8$\times$7B & Leaderboard & 47B MoE & 61.2 \\
GPT-5.4$\dagger$ & API & undiscl. & 62.5 \\
Claude Haiku$\dagger$ & API & undiscl. & 63.1 \\
Claude Sonnet$\dagger$ & API & undiscl. & 64.0 \\
GPT-4o-mini & API & undiscl. & 62.4 \\
Qwen-1.5 72B & Leaderboard & 72B & 64.0 \\
GPT-4o & API & undiscl. & 63.3 \\
Mixtral 8$\times$22B & Leaderboard & 141B MoE & 70.4 \\
Llama-3 70B Inst. & Leaderboard & 70B & 76.8 \\
\bottomrule
\end{tabular}
\end{table}

% ============================================================================
\section{Dataset and Code Documentation}
\label{app:datadoc}
% ============================================================================

\textbf{What is released.} We release: (1)~the annotated failure corpus (2,664 tasks from 6 benchmarks, evaluated across 18 models with full coverage and 155 models via leaderboard data); (2)~the behavioral failure taxonomy (25 base clusters, 26 hierarchical sub-skills, with GPT-5.4 labels and cross-LLM validation data); (3)~1,203 generated adversarial and metamorphic evaluation tasks; (4)~all embeddings (semantic, behavioral, combined); (5)~scalar ranking and structured diagnosis comparison data; (6)~the multi-turn corpus (363 frontier-hard conversations $\times$ 5 frontier models with depth-conditioned pass/fail data) used in \S\ref{sec:multiturn}; (7)~the multi-turn adversarial agent corpus (630 traces across 3 attack families $\times$ 3 frontier models with judge labels and real-execution outcomes) used in \S\ref{sec:adversarial_case}; and (8)~the full pipeline code.

\textbf{Licensing.} Code is released under the MIT License; the three FailureScope-released datasets (single-turn annotations on top of source benchmarks, multi-turn corpus, adversarial corpus) are released under CC-BY-4.0. Task data inherits the licenses of its source benchmarks: GSM8K (MIT), ARC (CC-BY-SA), MMLU (MIT), HumanEval (MIT), MBPP (CC-BY-4.0), IFEval (Apache-2.0). Generated tasks are released under CC-BY-4.0.

\textbf{Hosting and maintenance.} Data and code live on GitHub. Each of the three corpora is pinned to its own permanent Zenodo DOI: single-turn \texttt{10.5281/zenodo.20034013}, multi-turn \texttt{10.5281/zenodo.20034373}, adversarial \texttt{10.5281/zenodo.20034377}. We commit to keeping the repository maintained for at least two years past publication.

\textbf{Independence from concurrent work.} Each of the three corpora is archived on its own and reproducible from the released code and raw eval traces, whatever happens to any related concurrent submission, accepted, rejected, or withdrawn. The FailureScope release bundles everything needed: replication artifacts, LOMO clustering scripts, the permutation-test code (\texttt{scripts/permutation\_silhouette.py}), embedding files, and the cluster-quality metrics in Table~\ref{tab:embedding_ablation}. None of it hinges on the acceptance status of another paper.

\textbf{Intended use.} The dataset is intended for LLM evaluation research and model development. It should not be used for training language models (benchmark contamination) or for constructing adversarial attacks against deployed systems.

\textbf{Personal data.} The corpus contains no personal, private, or sensitive information. All tasks are drawn from publicly available benchmarks.

\textbf{Consent and compensation.} No human subjects or crowdworkers were involved. All evaluations use publicly available benchmarks and model APIs.

% ============================================================================
\section{Pipeline Algorithm}
\label{app:algorithm}
% ============================================================================

\begin{algorithm}[h]
\caption{\textsc{FailureScope} Pipeline (Stages 1--6)}
\label{alg:pipeline}
\begin{algorithmic}[1]
\REQUIRE Benchmark configs $\mathcal{B}$, model set $\mathcal{M}$, frontier set $\mathcal{M}_f \subseteq \mathcal{M}$
\ENSURE Failure taxonomy $\mathcal{T}$, adversarial corpus $\mathcal{A}$

\STATE \textbf{// Stage 1: Collection}
\STATE $\mathcal{C} \leftarrow \emptyset$
\FOR{benchmark $b \in \mathcal{B}$}
    \STATE $\mathcal{C} \leftarrow \mathcal{C} \cup \textsc{LoadTasks}(b, N_{\max})$
\ENDFOR

\STATE \textbf{// Stage 2: Evaluation}
\STATE $\mathcal{R} \leftarrow \emptyset$
\FOR{model $m \in \mathcal{M}$, task $t \in \mathcal{C}$}
    \STATE $r \leftarrow \textsc{Evaluate}(m, t)$ \COMMENT{Tier-1 deterministic or Tier-2 LLM judge}
    \STATE $\mathcal{R} \leftarrow \mathcal{R} \cup \{r\}$
\ENDFOR

\STATE \textbf{// Stage 3: Filtering}
\STATE $\mathcal{F} \leftarrow \{t \in \mathcal{C} \mid \exists\, m \in \mathcal{M}_f : \neg r_{m,t}.\text{passed}\}$

\STATE \textbf{// Stage 4: Embedding}
\FOR{task $t \in \mathcal{F}$}
    \STATE $\mathbf{z}_{\text{sem}} \leftarrow \textsc{VoyageEmbed}(t.\text{text})$
    \STATE $\mathbf{z}_{\text{beh}} \leftarrow [\mathds{1}[r_{m,t}.\text{passed}]]_{m \in \mathcal{M}}$
    \STATE $\mathbf{z}_t \leftarrow [\alpha \cdot \hat{\mathbf{z}}_{\text{sem}} \| (1-\alpha) \cdot \hat{\mathbf{z}}_{\text{beh}}]$
\ENDFOR

\STATE \textbf{// Stage 5: Taxonomy}
\STATE $\mathbf{Z}_{\text{red}} \leftarrow \textsc{PCA}([\mathbf{z}_t]_{t \in \mathcal{F}}, d=30)$
\STATE $\ell \leftarrow \textsc{HDBSCAN}(\mathbf{Z}_{\text{red}})$
\FOR{cluster $c_k$ in unique$(\ell)$}
    \STATE exemplars $\leftarrow$ 5 nearest tasks to centroid of $c_k$
    \STATE $(c_k.\text{label}, c_k.\text{mode}, c_k.\text{bloom}) \leftarrow \textsc{LLMLabel}(\text{exemplars})$
\ENDFOR
\STATE $\mathcal{T} \leftarrow \{c_k\}$

\STATE \textbf{// Stage 6: Generation}
\STATE $\mathcal{A} \leftarrow \emptyset$
\FOR{cluster $c_k \in \mathcal{T}$}
    \STATE candidates $\leftarrow \textsc{GenerateAdversarial}(c_k, 50 \times 3)$
    \STATE validated $\leftarrow \textsc{Validate}(\text{candidates})$
    \STATE $\mathcal{A} \leftarrow \mathcal{A} \cup \textsc{FilterContamination}(\text{validated}, \mathcal{C})$
\ENDFOR

\RETURN $\mathcal{T}, \mathcal{A}$
\end{algorithmic}
\end{algorithm}

\begin{algorithm}[h]
\caption{\textsc{FailureScope} Cross-Model Prediction (Stage 7)}
\label{alg:crossmodel}
\begin{algorithmic}[1]
\REQUIRE Taxonomy $\mathcal{T}$, model set $\mathcal{M}$, results $\mathcal{R}$
\ENSURE Per-model prediction AUCs

\FOR{held-out model $m^* \in \mathcal{M}$}
    \STATE $\mathcal{M}_{\text{train}} \leftarrow \mathcal{M} \setminus \{m^*\}$
    \STATE Rebuild taxonomy $\mathcal{T}'$ using $\mathcal{M}_{\text{train}}$
    \FOR{cluster $c_k \in \mathcal{T}'$}
        \STATE $\hat{f}_{m^*,k} \leftarrow \textsc{PredictFailure}(m^*, c_k, \mathcal{M}_{\text{train}}, \mathcal{R})$
        \STATE $f_{m^*,k} \leftarrow \text{actual failure rate of } m^* \text{ on } c_k$
    \ENDFOR
    \STATE $\text{AUC}_{m^*} \leftarrow \textsc{ComputeAUC}(\hat{f}_{m^*}, f_{m^*}, \theta)$
\ENDFOR
\end{algorithmic}
\end{algorithm}

% ============================================================================
\section{HDBSCAN Sensitivity Analysis}
\label{app:hdbscan}
% ============================================================================

\begin{table}[t]
\centering
\caption{Effect of \texttt{min\_cluster\_size} on the \emph{base} HDBSCAN partition, prior to mega-cluster sub-clustering. Cluster counts here are the first-stage HDBSCAN clusters; the two-stage sub-clustering step (Section~\ref{sec:method}) then refines this base partition into the final taxonomy reported in the main results.}
\small
\begin{tabular}{cccc}
\toprule
\texttt{min\_cluster\_size} & \textbf{Clusters} & \textbf{Noise \%} & \textbf{Silhouette} \\
\midrule
5 & 10 & 0.0\% & 0.073 \\
8 & 9 & 0.0\% & 0.082 \\
10 & 9 & 0.0\% & 0.082 \\
15 & 8 & 0.0\% & 0.095 \\
20 & 8 & 0.0\% & 0.095 \\
25 & 8 & 0.0\% & 0.095 \\
30 & 7 & 1.9\% & 0.095 \\
\bottomrule
\end{tabular}
\end{table}

These rows report the \emph{base} (first-stage) HDBSCAN partition before mega-cluster sub-clustering; our final taxonomy applies the two-stage sub-clustering of Section~\ref{sec:method} on top of this base partition, and the silhouette ($0.119$) and noise ($3.0\%$) reported in the main results are for that final taxonomy. The base partition is stable across \texttt{min\_cluster\_size}: base-cluster count varies only from 7 to 10 across a $6\times$ range of the hyperparameter, and base silhouette scores vary from 0.073 to 0.095. All configurations produce zero noise points except \texttt{min\_cluster\_size}$=$30 (1.9\% noise). We use 10 as the default as a conservative choice.

% ============================================================================
\section{Failure Mode Definitions}
\label{app:failure_modes}
% ============================================================================

\begin{table}[t]
\centering
\caption{Failure mode taxonomy with definitions and examples.}
\small
\begin{tabular}{p{3.5cm}p{5.5cm}p{4.5cm}}
\toprule
\textbf{Failure Mode} & \textbf{Definition} & \textbf{Example} \\
\midrule
Knowledge gap & Model lacks the factual or domain knowledge required & Obscure historical dates, rare scientific constants \\
Reasoning failure & Model has the knowledge but composes it incorrectly & Multi-step math errors, invalid logical inferences \\
Instruction drift & Model ignores or gradually deviates from task constraints & Violating format requirements, exceeding length limits \\
Brittleness & Small surface-level changes cause failure on previously-solved tasks & Rephrased questions, reordered premises \\
Hallucination & Model confidently produces false information & Fabricated citations, invented facts \\
Edge-case blindness & Model fails on boundary conditions or unusual inputs & Division by zero, empty input handling \\
\bottomrule
\end{tabular}
\end{table}

% ============================================================================
\section{Per-Model Failure Profiles}
\label{app:profiles}
% ============================================================================

\begin{table}[t]
\centering
\caption{Per-model failure rates across representative failure categories. ``---'' indicates the model was not evaluated on tasks in that cluster.}
\label{tab:model_profiles}
\small
\begin{tabular}{lcccc|cccc}
\toprule
& \multicolumn{4}{c|}{\textbf{API Models}} & \multicolumn{4}{c}{\textbf{Open Models}} \\
\textbf{Category} & \textbf{5.4} & \textbf{5.4-n} & \textbf{Son.} & \textbf{Hai.} & \textbf{L2-7} & \textbf{Mis-7} & \textbf{L2-70} & \textbf{L3-70} \\
\midrule
Python Func.\ Impl. & .96 & .95 & .95 & .94 & --- & --- & --- & --- \\
Arith.\ Reasoning & .28 & .26 & .24 & .25 & .96 & .87 & .77 & .23 \\
Scientific Reasoning & .02 & .02 & .02 & .02 & .88 & .85 & .78 & .73 \\
Astronomy Gaps & .94 & .96 & .91 & .93 & --- & --- & --- & --- \\
Anatomy Knowledge & .92 & .89 & .94 & .93 & 1.0 & .50 & .50 & .50 \\
Abstract Algebra & .94 & .90 & .93 & .93 & .67 & .83 & .50 & .50 \\
Business Ethics & .97 & .96 & .93 & .93 & --- & --- & --- & --- \\
Bus.\ Governance & .96 & 1.0 & 1.0 & .93 & .52 & .33 & .22 & .07 \\
Age Word Problems & .00 & .00 & .10 & .20 & 1.0 & 1.0 & .80 & .40 \\
\midrule
\textbf{Overall} & \textbf{.80} & \textbf{.87} & \textbf{.77} & \textbf{.78} & \textbf{.90} & \textbf{.83} & \textbf{.74} & \textbf{.44} \\
\bottomrule
\end{tabular}
\end{table}

Notable patterns: API models (GPT-5.4, Claude Sonnet/Haiku) show high failure rates on frontier-hard tasks, ranging from 77\% (Claude Sonnet) to 87\% (GPT-5.4-nano). GPT-5.4-nano shows distinctly higher failure rates than the other frontier models. Variance among open models runs much wider: Llama-2 7B fails on 90\% of tasks, Llama-3 70B Instruct on only 44\%. The ``Age-Related Word Problems'' cluster shows the starkest model-size effect: all API models achieve near-perfect accuracy (0--20\% failure) while smaller open models fail completely (100\%). The gap points to a capability that scales sharply with model size.

% ============================================================================
\section{Oracle Upper Bound (All-Model Taxonomy)}
\label{app:oracle}
% ============================================================================

\begin{table}[t]
\centering
\caption{Oracle diagnostic efficiency: taxonomy built from all 18 models (including evaluation targets). Mean Kendall's $\tau$ with 95\% CIs across models ($n{=}18$, 100 trials each). These numbers represent the theoretical ceiling of taxonomy-based diagnosis.}
\label{tab:oracle}
\small
\begin{tabular}{lcccccc}
\toprule
\textbf{Strategy} & $N{=}10$ & $N{=}25$ & $N{=}50$ & $N{=}100$ & $N{=}250$ & $N{=}500$ \\
\midrule
Random & .01$\pm$.01 & .13$\pm$.09 & .47$\pm$.12 & .84$\pm$.13 & .91$\pm$.09 & .94$\pm$.08 \\
Benchmark stratified & .02$\pm$.02 & .25$\pm$.20 & .52$\pm$.13 & .86$\pm$.12 & .93$\pm$.07 & .95$\pm$.06 \\
Taxonomy conditioned & \textbf{.94}$\pm$.13 & \textbf{.95}$\pm$.11 & .88$\pm$.13 & .90$\pm$.11 & .94$\pm$.08 & \textbf{.96}$\pm$.04 \\
Difficulty stratified & .01$\pm$.01 & .07$\pm$.06 & .29$\pm$.13 & .69$\pm$.11 & .91$\pm$.09 & .94$\pm$.06 \\
IRT-adaptive & .00$\pm$.00 & .56$\pm$.21 & .82$\pm$.17 & .92$\pm$.09 & \textbf{.97}$\pm$.04 & \textbf{.99}$\pm$.02 \\
Uncertainty & .01$\pm$.01 & .41$\pm$.15 & .80$\pm$.19 & .90$\pm$.11 & .95$\pm$.06 & .98$\pm$.03 \\
\bottomrule
\end{tabular}
\end{table}

The oracle taxonomy-conditioned strategy achieves $\tau = 0.94$ at just $N{=}10$ tasks, demonstrating the theoretical ceiling when the taxonomy perfectly captures the evaluation targets' failure patterns. Comparing with LOMO results (Table~\ref{tab:efficiency}), the gap narrows from $0.61$ at $N{=}10$ (oracle $0.94$ vs.\ LOMO $0.33$) to $0.14$ at $N{=}500$ (oracle $0.96$ vs.\ LOMO $0.82$), indicating that taxonomy generalization improves with sample size.

% ============================================================================
\section{Allocation Ablation}
\label{app:allocation}
% ============================================================================

\begin{table}[t]
\centering
\caption{Proportional vs.\ uniform cluster allocation for taxonomy-conditioned sampling (oracle taxonomy). Mean Kendall's $\tau$ across 18 models, 100 trials.}
\label{tab:allocation_ablation}
\small
\begin{tabular}{lcccccc}
\toprule
\textbf{Allocation} & $N{=}10$ & $N{=}25$ & $N{=}50$ & $N{=}100$ & $N{=}250$ & $N{=}500$ \\
\midrule
Proportional & \textbf{.94} & \textbf{.95} & .88 & .90 & .93 & .96 \\
Uniform & .00 & .00 & \textbf{.88} & \textbf{.92} & \textbf{.97} & \textbf{.99} \\
\bottomrule
\end{tabular}
\end{table}

At low budgets ($N \leq 25$), proportional allocation is essential: with 25 clusters and $N{=}25$, uniform allocation assigns $\sim$1.0 tasks per cluster, below the $k_{\min}{=}3$ threshold needed for reliable failure-rate estimation. Proportional allocation guarantees minimum coverage per cluster first, then distributes the remaining budget proportional to cluster size. At higher budgets ($N \geq 50$), the difference narrows as both strategies provide sufficient per-cluster coverage. Uniform allocation slightly outperforms at $N \geq 100$ because equal representation avoids under-sampling small clusters.

% ============================================================================
\section{Imputation Method Comparison}
\label{app:imputation}
% ============================================================================

\begin{table}[t]
\centering
\caption{Imputation method comparison for behavioral matrix (55.2\% missing). All methods use $\alpha{=}0.1$ (behavioral-dominant) with HDBSCAN clustering.}
\label{tab:imputation}
\small
\begin{tabular}{lcccc}
\toprule
\textbf{Method} & \textbf{Silhouette} & \textbf{Clusters} & \textbf{Noise (\%)} & \textbf{Cross-bench (\%)} \\
\midrule
Column mean (ours) & 0.119 & 25 & 3.0 & 86.7 \\
Zero fill & 0.774 & 10 & 0.9 & 90.0 \\
Half fill (0.5) & 0.680 & 14 & 1.0 & 85.7 \\
Row mean & 0.775 & 10 & 0.9 & 90.0 \\
No imputation & 0.562 & 13 & 1.0 & 84.6 \\
\bottomrule
\end{tabular}
\end{table}

Column-mean imputation (our choice) provides a good balance of cluster resolution (25 clusters) and cross-benchmark representation (86.7\%), with silhouette 0.119 and only 3.0\% noise. Imputation substantially improves clustering quality compared to no imputation, confirming that the missing entries in the raw behavioral matrix distort the embedding geometry.

% ============================================================================
\section{Cross-LLM Label Validation}
\label{app:label_validation}
% ============================================================================

To assess the robustness of our GPT-5.4 cluster labels, we independently re-label the clusters of an earlier 15-cluster taxonomy build using GPT-5.4-nano with identical prompts and exemplars; agreement is reported over those 15 clusters. Table~\ref{tab:label_validation} summarizes agreement metrics.

\begin{table}[t]
\centering
\caption{Cross-LLM label validation: GPT-5.4 (original) vs.\ GPT-5.4-nano (validator). Agreement is measured across 15 clusters. Task membership: fraction of sampled tasks (10 per cluster) that GPT-5.4-nano judges as matching the cluster's original label.}
\label{tab:label_validation}
\small
\begin{tabular}{lcc}
\toprule
\textbf{Metric} & \textbf{Value} & \textbf{Interpretation} \\
\midrule
Failure mode exact agreement & 73.3\% & Majority of 15 clusters agree \\
Failure mode Cohen's $\kappa$ & 0.250 & Fair agreement \\
Bloom level exact agreement & 66.7\% & Majority of 15 clusters agree \\
Bloom level Cohen's $\kappa$ & 0.479 & Moderate agreement \\
Task membership (overall) & 54.7\% & Sampled tasks match \\
Task membership (arith.\ clusters) & 93.3\% & High agreement \\
Task membership (mega-cluster subs) & 30.0\% & Heterogeneous clusters \\
\bottomrule
\end{tabular}
\end{table}

The four failure-mode disagreements all involve GPT-5.4 labeling clusters as \emph{reasoning\_failure} where GPT-5.4-nano labels them as \emph{knowledge\_gap}. This boundary is inherently ambiguous: a task requiring multi-step scientific reasoning could be categorized as either a reasoning failure (the model cannot compose the steps) or a knowledge gap (the model lacks the domain knowledge to reason about). The moderate kappa scores are consistent with the inter-annotator agreement literature for subjective categorization tasks~\citep{ribeiro2020checklist}. Clusters with clear, homogeneous failure patterns (e.g., ``Complex Multi-step Arithmetic Problems'') achieve near-perfect membership validation (80--100\%), while heterogeneous or semantically broad clusters score lower.

% ============================================================================
\section{Scalar Ranking vs.\ Structured Diagnosis}
\label{app:scalar_ranking}
% ============================================================================

Table~\ref{tab:scalar_vs_structured} directly contrasts the two evaluation objectives at $N{=}25$. Spearman's $\rho$ measures how well the strategy recovers the true \emph{overall model accuracy ranking}; Kendall's $\tau$ measures how well it recovers the true \emph{per-cluster weakness ranking}.

\begin{table}[t]
\centering
\caption{Head-to-head comparison at $N{=}25$: scalar ranking (Spearman $\rho$, higher is better) vs.\ structured diagnosis (Kendall $\tau$, higher is better). No single strategy dominates both objectives.}
\label{tab:scalar_vs_structured}
\small
\begin{tabular}{lcc}
\toprule
\textbf{Strategy} & \textbf{Scalar $\rho$ ($N{=}25$)} & \textbf{Diagnosis $\tau$ ($N{=}25$)} \\
\midrule
Random & 0.72 & 0.31 \\
Benchmark-stratified & 0.75 & 0.24 \\
Difficulty-stratified & 0.67 & 0.31 \\
IRT-adaptive & \textbf{0.77} & 0.00 \\
Uncertainty & 0.76 & 0.28 \\
Taxonomy-conditioned & 0.60 & \textbf{0.33} \\
\bottomrule
\end{tabular}
\end{table}

% ============================================================================
% NeurIPS Paper Checklist
% ============================================================================

\section*{NeurIPS Paper Checklist}

\begin{enumerate}

\item {\bf Claims}
    \item[] Question: Do the main claims made in the abstract and introduction accurately reflect the paper's contributions and scope?
    \item[] Answer: \answerYes{}
    \item[] Justification: The abstract states taxonomy-conditioned sampling achieves $\tau = 0.81$ at 50 tasks, validated via LOMO protocol across 18 models. These claims are supported by Table~\ref{tab:efficiency} and Section~\ref{sec:rq2}; cross-model failure prediction (AUC 0.88) by Section~\ref{sec:rq3}; the six-cluster multi-turn taxonomy by Section~\ref{sec:multiturn}; and the 73--100 percentage-point judge-vs-execution gap by Section~\ref{sec:adversarial_case}.

\item {\bf Limitations}
    \item[] Question: Does the paper discuss the limitations of the work performed by the authors?
    \item[] Answer: \answerYes{}
    \item[] Justification: Section~\ref{sec:discussion} summarizes limitations and Appendix~\ref{app:limitations_full} enumerates eleven items, including model pool scale, extrapolation concerns, data sparsity, LLM labeling subjectivity, binary pass/fail encoding, taxonomy construction cost, benchmark coverage, multi-turn and adversarial model pools, the generator-judge confound, and mega-cluster granularity.

\item {\bf Theory Assumptions and Proofs}
    \item[] Question: For each theoretical result, does the paper provide the full set of assumptions and a complete (correct) proof?
    \item[] Answer: \answerNA{}
    \item[] Justification: The paper is empirical and does not make theoretical claims requiring formal proofs.

\item {\bf Experiments Reproducibility}
    \item[] Question: Does the paper fully disclose all the information needed to reproduce the main experimental results of the paper to the extent that it affects the main claims and/or conclusions of the paper?
    \item[] Answer: \answerYes{}
    \item[] Justification: Section~\ref{sec:method} specifies all hyperparameters (HDBSCAN settings, $\alpha$, PCA components, imputation method). The pipeline algorithm is provided in Appendix~\ref{app:algorithm}. Code and data are released.

\item {\bf Open access to data and code}
    \item[] Question: Does the paper provide open access to the data and code, with sufficient instructions to faithfully reproduce the main experimental results?
    \item[] Answer: \answerYes{}
    \item[] Justification: The Dataset and Code Documentation section describes all released artifacts: annotated failure corpus, taxonomy, generated tasks, embeddings, and pipeline code under MIT license, with Zenodo archival.

\item {\bf Experimental Setting/Details}
    \item[] Question: Does the paper specify all the training and test details necessary to understand the results?
    \item[] Answer: \answerYes{}
    \item[] Justification: Section~\ref{sec:setup} specifies the corpus (2,664 tasks, 6 benchmarks), model set (18 models, Table~\ref{tab:models}), evaluation protocol (LOMO), baselines, and all metrics. Appendix sections provide HDBSCAN sensitivity, imputation comparisons, and allocation ablations.

\item {\bf Experiment Statistical Significance}
    \item[] Question: Does the paper report error bars suitably and correctly defined or other appropriate information about the statistical significance of the experiments?
    \item[] Answer: \answerYes{}
    \item[] Justification: Table~\ref{tab:efficiency} reports 95\% confidence intervals across 18 model folds and 50 trials each. Figure~\ref{fig:efficiency} includes error bars.

\item {\bf Experiments Compute Resources}
    \item[] Question: For each experiment, does the paper provide sufficient information on the computer resources needed to reproduce the experiments?
    \item[] Answer: \answerPartially{}
    \item[] Justification: The paper describes API-based evaluation (asynchronous calls with concurrency limit of 10) and specifies which models are evaluated via API vs.\ leaderboard data. Exact API costs are not reported.

\item {\bf Code Of Ethics}
    \item[] Question: Does the research conducted in the paper conform, in every respect, with the NeurIPS Code of Ethics?
    \item[] Answer: \answerYes{}
    \item[] Justification: The research uses publicly available benchmarks and model APIs. No human subjects are involved. The Broader Impact section discusses potential risks and mitigations.

\item {\bf Broader Impacts}
    \item[] Question: Does the paper discuss both potential positive societal impacts and negative societal impacts of the work?
    \item[] Answer: \answerYes{}
    \item[] Justification: The Broader Impact section discusses positive applications (efficient evaluation, failure transparency) and potential risks (adversarial exploitation, benchmark bias, label bias).

\item {\bf Safeguards}
    \item[] Question: Does the paper describe safeguards that have been put in place for responsible release of data or models that have a high risk for misuse?
    \item[] Answer: \answerYes{}
    \item[] Justification: The Broader Impact section notes that adversarial generation focuses on standard evaluation tasks rather than safety-critical domains. The Dataset and Code Documentation section specifies intended use and warns against training contamination.

\item {\bf Licenses for existing assets}
    \item[] Question: Are the creators or original owners of assets used in the paper properly credited and are the license and terms of use explicitly mentioned and properly respected?
    \item[] Answer: \answerYes{}
    \item[] Justification: The Dataset and Code Documentation section lists licenses for all source benchmarks (GSM8K, ARC, MMLU, HumanEval, MBPP, IFEval) and specifies CC-BY-4.0 for generated tasks.

\item {\bf New Assets}
    \item[] Question: Are new assets introduced in the paper well documented and is the documentation provided alongside the assets in a dataset card?
    \item[] Answer: \answerYes{}
    \item[] Justification: The Dataset and Code Documentation section describes all released artifacts, licensing, hosting, intended use, and maintenance commitment.

\item {\bf Crowdsourcing and Research with Human Subjects}
    \item[] Question: For crowdsourcing experiments and research with human subjects, does the paper include the full text of instructions given to participants and screenshots, if applicable?
    \item[] Answer: \answerNA{}
    \item[] Justification: No crowdsourcing or human subjects research was conducted. All evaluations use automated pipelines and publicly available data.

\item {\bf Institutional Review Board (IRB) Approvals or Equivalent for Research with Human Subjects}
    \item[] Question: Does the paper describe potential risks incurred by study participants, whether compensation was provided, and whether informed consent was obtained?
    \item[] Answer: \answerNA{}
    \item[] Justification: No human subjects were involved in this research.

\end{enumerate}

\end{document}